\documentclass{bioinfo}

\copyrightyear{2018} \pubyear{2018}
\usepackage{multirow}
\usepackage[table, x11names]{xcolor}
\usepackage{listings}
\usepackage{color}
\access{Advance Access Publication Date: Day Month Year}
\appnotes{Manuscript Category}

\usepackage{graphicx}
\usepackage{epstopdf}
\usepackage{amsmath}
\usepackage{caption}
\usepackage{hyperref}

\begin{document}
\firstpage{1}

\subtitle{Data and text mining}

\title{EBIC: an evolutionary-based parallel biclustering algorithm for pattern discovery}

\author{Patryk Orzechowski\,$^{1,2}$, Moshe Sipper\,$^{1,3}$, Xiuzhen Huang\,$^4$ \& Jason H. Moore\,$^{*1}$}

\author[Orzechowski \textit{et~al}.]{Patryk Orzechowski\,$^{\text{\sfb 1,}\text{\sfb 2} *}$, Moshe Sipper\,$^{\text{\sfb 3}}$, Xiuzhen Huang\,$^{\text{\sfb 4}}$, and\\Jason H. Moore\,$^{\text{\sfb 1}*}$}
\address{
$^{\text{\sf 1}}$Institute for Biomedical Informatics, University of Pennsylvania, Philadelphia, PA 19104, USA, and \\
$^{\text{\sf 2}}$Department of Automatics and Biomedical Engineering, AGH University of Science and Technology, al. Mickiewicza 30, 30-059 Krakow, Poland, and \\
$^{\text{\sf 3}}$Department of Computer Science, Ben-Gurion University, Beer Sheva 8410501, Israel,and \\
$^{\text{\sf 4}}$Department of Computer Science, Arkansas State University, Jonesboro, AR 72467, USA
}

\corresp{$^\ast$To whom correspondence should be addressed.}

\history{Received on XXXXX; revised on XXXXX; accepted on XXXXX}

\editor{Associate Editor: XXXXXXX}


\abstract{
\textbf{Motivation: } Biclustering algorithms are commonly used for gene expression data analysis. However, accurate identification of meaningful structures is very challenging and state-of-the-art methods are incapable of discovering with high accuracy different patterns of high biological relevance. \\
\textbf{Results: }In this paper a novel biclustering algorithm based on evolutionary computation, a subfield of artificial intelligence (AI), is introduced. The method called EBIC aims to detect order-preserving patterns in complex data. EBIC is capable of discovering multiple complex patterns with unprecedented accuracy in real gene expression datasets. It is also one of the very few biclustering methods designed for parallel environments with multiple graphics processing units (GPUs). We demonstrate that EBIC greatly outperforms state-of-the-art biclustering methods, in terms of recovery and relevance, on both synthetic and genetic datasets. EBIC also yields results over 12 times faster than the most accurate reference algorithms.
\\
\textbf{Availability: } EBIC source code is available on GitHub at \url{https://github.com/EpistasisLab/ebic}\\
\textbf{Contact: }Correspondence and requests for materials should be addressed to P.O. ~(email: patryk.orzechowski@gmail.com) and J.H.M. ~(email: jhmoore@upenn.edu)\\
\textbf{Supplementary information: }Supplementary Data with results of analyses and additional information on the method is available at \emph{Bioinformatics} online.}

\maketitle

\section{Introduction}

Discovering meaningful patterns in complex and noisy data, especially biological one, is a challenge. Traditional clustering approaches such as k-means or hierarchical clustering are expected to group similar objects together and to separate dissimilar objects into distinctive groups. These methods assume that all object features contribute to the classification result, which renders clustering a valuable technique for global similarity detection. Clustering does not, however, succeed when only some subset of features is important to a specific cluster. 

The inability to capture local patterns is one of the main reasons for the advent of biclustering techniques, where biclusters -- subsets of rows and columns -- are sought.  Both rows and columns subsets may contain elements that are not necessarily adjacent to each other, thus differentiating biclustering from other problems of pattern matching (e.g., image recognition), making also the task unsuitable for deep learning  \citep{Ching2017}. 

Biclustering has its roots in data partitioning into subgroups of approximately constant values \citep{Morgan1963} and simultaneously clustering rows and columns of a matrix \citep{Hartigan1972}; this was later called biclustering  \citep{Mirkin1996}. For the last two decades biclustering has been applied to multiple domains, including biomedicine, genomics (especially gene expression analysis), text-mining, marketing, dimensionality reduction, and others  \citep{Busygin2008,Dolnicar2012}.  

Designing biclustering algorithms involves many challenges. First, although over fifty biclustering algorithms have been proposed (much more when derivatives are considered), no method has proven capable of detecting  --  with sufficient accuracy  --  six major types of patterns that are commonly present in gene expression data. Most biclustering algorithms find only one or a few of these patterns  \citep{Madeira2004,Eren2013,Pontes2015,Wang2016,Padilha2017}: column-constant, row-constant, shift (i.e., additive coherent), scale (i.e., multiplicative coherent), shift and scale (i.e., simultaneous coherent), and order-preserving. Detection of order-preserving patterns is especially important, because it may be considered a generalization of the five other patterns \citep{BenDor2003,Eren2013,Wang2016}.  

Second, many biclustering algorithms are unable to detect negative correlations or capture approximate patterns. Moreover, biclustering algorithms fail to properly separate partially overlapping biclusters. The performance of these algorithms on overlapping problems usually drops dramatically with increasing levels of overlap \citep{Wang2016}. 

A third drawback of current biclustering methods is their limited success assessing which biclusters are the most relevant. Multiple measures for assessing quality of biclusters have been used so far \citep{Orzechowski2013,Pontes2015quality}. Some algorithms yield only a single bicluster at a time, rendering their application cumbersome \citep{Pontes2015}. Other methods output a high number of biclusters (e.g., BiMax \citep{Prelic2006} and PBBA \citep{Orzechowski2016}). This usually produces many overlaps and degrades the overall performance of the algorithm \citep{Eren2013}. 

Providing the proper balance between local and global context within the data is also difficult. The methods that model global relations are typically able to deliver only a limited number of results  (e.g., Plaid \citep{Lazzeroni2002}, FABIA \citep{Hochreiter2010}, and ISA \citep{Bergmann2003}), or tend to exhibit decreased accuracy with each result (e.g., CC \citep{Cheng2000}). On the other hand, algorithms that focus on local similarities are susceptible to losing global reference (e.g., Bimax, PBBA, or UniBic \citep{Wang2016}). For example, UniBic, which sorts pairs of values and column indices of each row in order to identify the longest common subsequences, is able to detect the longest order-preserving pattern between each pair of rows, irrespective of the order of columns, but it fails to capture narrow biclusters containing only a few rows and multiple columns.

As the biclustering problem is NP-hard, designing an efficient and accurate parallel biclustering algorithm remains a challenge. Most of the reference biclustering algorithms are purely sequential. The reason for this is that the methods either require intensive computations, which limit their application to datasets of smaller size, or are fast but at the cost of lower accuracy.


\section{Methods}
In this paper a novel biclustering algorithm called EBIC is introduced, which overcomes the above shortcomings. 
The algorithm is based on evolutionary computation, a subfield of Artificial Intelligence (AI). It is likely the first biclustering algorithm capable of detecting all aforementioned types of meaningful patterns with very high accuracy. EBIC is also one of very few parallel biclustering methods. We show that the proposed algorithm outperforms the most established methods in the field with respect to accuracy and relevance on both synthetic and real genomic datasets. An open-source, multi-GPU, parallel implementation of the algorithm is also provided.

The algorithm is designed for environments with at least a single GPU and requires the installation of CUDA. The algorithm was developed in C++11 with OpenMP, with CUDA used for parallelization. 

\subsection{Motivation.}
The design of the algorithm is motivated by the following observation. Given the input matrix $A=\{a_{ij}\}$, where $i$ stands for rows and $j$ for columns, consider counting the number of rows with the property that the value in column $p$ is smaller than the value in column $q$, i.e. $\#\{k: a_{kp}<a_{kq}\}$. If the values in the dataset are generated randomly with univariate distribution, half of the rows on average are expected to have this property, and half are not. Addition of another column $r$ to the series, such that values in this columns are larger than the values in column $q$, i.e. $\# \{k: a_{kp}<a_{kq}<a_{kr}\}$, should result in another reduction of the number of rows by half. Thus, for data without any signal, each addition of a column to the series reduces the number of concordant rows by half. On the other hand, if the distribution of the data is not uniform and there exists a monotonic relationships between rows in some subset of conditions, any addition of the pattern-specific column won't eliminate the rows belonging to this pattern. Thus, the algorithm attempts to intelligently manipulate multiple series of columns and assigns higher scores to those series in which column additions do not result in total reduction of the rows.

The quality of each bicluster is determined by a function (called fitness), which takes into consideration the number of columns and, exponentially, the number of rows that follow the monotonically increasing trend represented by each series of columns. The design of the fitness function promotes incorporation of new columns to biclusters, provided there is a sufficient number of rows matching the trend (\ref{eq:fitness}).

\begin{equation} 
\label{eq:fitness}
f(B)=\begin{cases}
2^{\min(|I|-\sigma,0)} \cdot |J| \cdot \log( \max(|I|-1,0)) & if\, |I|>1 \\
0 & if\, |I| \leq 1 \,,
\end{cases}
\end{equation}
where $\sigma$ is the expected minimal number of rows that should be included within a bicluster $B=(I,J)$, with its rows and columns denoted as $I$ and $J$, respectively.

EBIC uses a different representation compared with other evolutionary-based biclustering methods \citep{Divina2006,Mitra2006,Ayadi2012}. Instead of modeling a bicluster as a tuple with a set of rows and a set of columns, biclusters in EBIC are represented by a series of column indices. The quality of a given series is calculated based on the number of rows that match the monotonous rules present within the series of columns. The modification of column series is performed using an AI-based technique known as genetic programming (GP) \citep{Koza1992,Poli2008}. Series of columns are expanded only when the rule they impose is matched by sufficient number of rows.

\begin{figure}[ht]
\begin{center}
\includegraphics[width=0.5\textwidth]{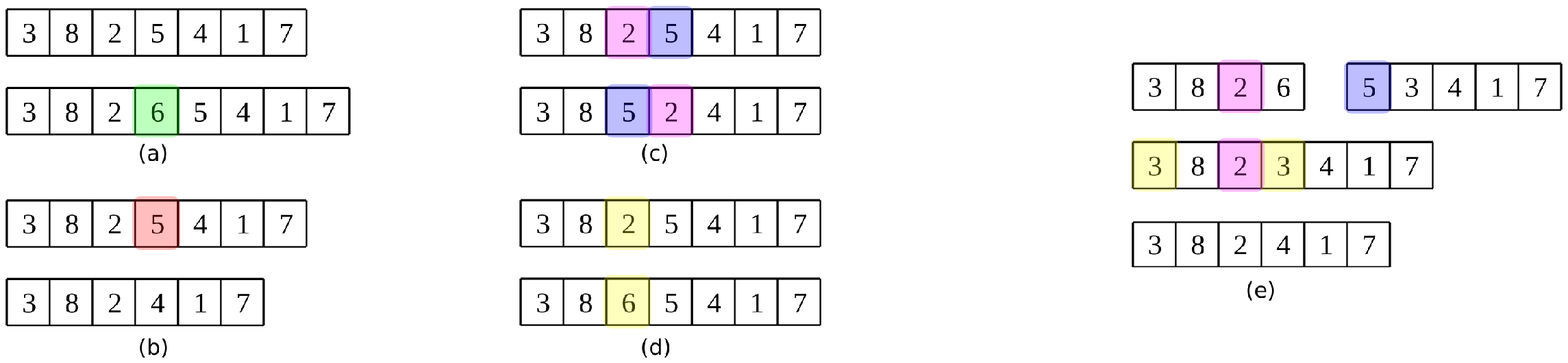}
\end{center}
\caption[Genetic operations in EBIC]{Genetic operations in EBIC: (a) insertion mutation, (b) deletion mutation, (c) swap mutation, (d) substitution mutation, and (e) crossover.}
\label{genetic-operations}
\end{figure}

EBIC belongs to the family of hybrid biclustering approaches \citep{Orzechowski2016hybrid} and features several techniques commonly used in evolutionary algorithms. The development of biclusters is driven by simple genetic operations: 1) four different types of mutations -- insertion of a new column to the series (Fig. \ref{genetic-operations}a), deletion of one of the columns from the series (Fig. \ref{genetic-operations}b), swap of two columns within the series (Fig. \ref{genetic-operations}c), and substitution of a column within the series (Fig. \ref{genetic-operations}d); and 2) crossover (Fig. \ref{genetic-operations}e).  The individuals that are set to undergo genetic operations are determined using tournament selection. To obtain a diverse set of solutions, a variant of a technique called \textit{crowding} is used, which limits the probability of selecting those individuals that share columns with those already added to the new generation \citep{Sareni1998}. More specifically, the fitness of individuals that take part in a tournament is decreased by the homogeneous penalty of $1.2^\vartheta$, where $\vartheta$ corresponds to the average penalty of using each of the columns separately. The explanation for this value of the parameter is provided in Supplementary Data. The described penalty enhances additions to the population individuals with underrepresented columns, what highly increases the diversity in population.

\begin{figure}[ht!]
\includegraphics[width=0.4\textwidth]{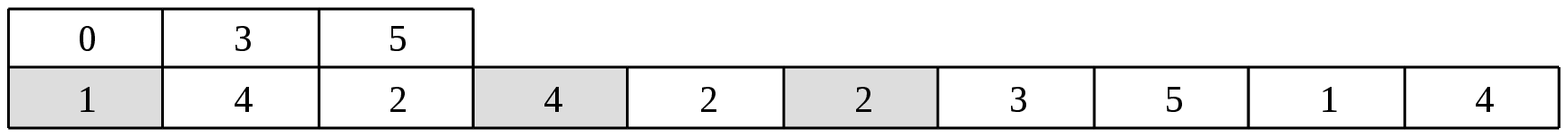}
\caption[Compressed Bicluster Format]{Compressed Bicluster Format (CBF) uses two arrays. The first array determines the starting positions of each of the biclusters, while the second one holds indexes of columns of biclusters. In this example the population consists of three biclusters (individuals): {(1,4,2), (4,2), and (2,3,5,1,4), which start at indices 0, 3, and 5, respectively.}}
\label{biclusters-cbf}
\end{figure}

Individuals whose overall fitness is the highest are stored in the top-rank list, which is updated only if a newly found individual does not substantially overlap with an individual in the list. During the construction of a new population a variant of a \textit{tabu list} is used, which forbids calculation of the previous biclusters \citep{Glover1989,Glover1990}. Elitism is used to clone a group of the best individuals found so far, so that the population is still able to search around local minima \citep{Poli2008elitism}. 
To limit the communication overhead, a Compressed Biclusters Format (CBF) is proposed for storing biclusters (see Fig. \ref{biclusters-cbf}). The format was motivated by Compressed Row Storage (CRS), a popular representation of sparse matrices. 

\subsection{EBIC Algorithm.} 
\begin{figure}[ht]
\begin{center}
\includegraphics[width=0.4\textwidth]{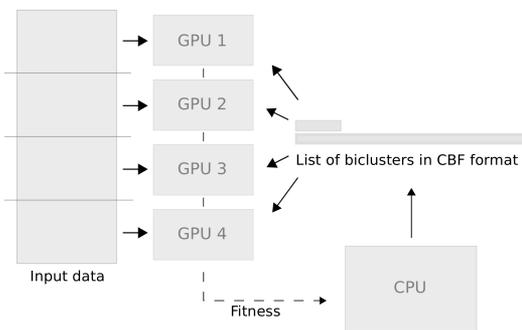}
\end{center}
\caption[Overview of EBIC]{Overview of EBIC. After dispatching chunks of the input data to multiple GPUs, biclusters -- represented by multiple series of columns and stored in CBF format -- are broadcast to GPUs. Each GPU calculates the number of how many rows of the chunk match the series imposed by the columns. This is used to determine fitness of each bicluster and generate a new set of biclusters. }
\label{algorithm}
\end{figure}

The basic concept of EBIC  -- a parallel biclustering algorithm based on Artificial Intelligence (AI) -- is presented in Figure \ref{algorithm}. 
The dataset is split into equal chunks of data and distributed across multiple GPUs. A population of different series of columns is generated on the CPU, stored in CBF format, and broadcast to multiple GPUs. Each GPU counts the number of rows which match the given series. The results are summarized on each GPU and sent back to the CPU in order to calculate fitness, which is used later to assess bicluster quality.

\paragraph{Step 1: Initialization.} Set up GPUs, divide the dataset proportionally by rows depending on the number of GPUs, and distribute the data across multiple GPUs. Generate initial population, calculate fitness on GPUs. Initialize top-rank list by sequentially adding unique (non-overlapping) series of columns with the highest fitness according to (\ref{eq:fitness}).

\paragraph{Step 2: Elitism.} Reproduce 1/4 of the best biclusters from the top-rank list, add them to the new population. Update penalties for using each column (each column addition to the population increases the penalty for using this column).

\paragraph{Step 3: Prepare population of biclusters.} Until the population reaches its required size, try to generate unique solutions (i.e., that haven't been previously analyzed). Select each new individual using tournament selection. Thus, select a solution randomly from the previous population and adjust its quality by applying the penalty for similarity with the previously accepted solutions. The penalty $vartheta$ is calculated by averaging penalties incurred by selecting each column separately over the number of columns within the series. The final penalty is calculated using the value of $1.2^\vartheta$. After selecting individuals, perform genetic operations (crossover and mutation). If the solution is novel (i.e., does not belong to the tabu list) add it to the population and the tabu list, and update penalties for using the solution's columns. Store the population in Compressed Biclusters Format (CBF). If the solution was previously analyzed, increase the number of tabu-list hits. If this number is greater than the size of population, finish calculations and go to step 6 in order to report the previously found best patterns. 

\paragraph{Step 4: Calculate quality of biclusters in parallel.} Dispatch the new population (i.e., sets of column series) to each of the GPUs. Determine how many rows match each of the series of columns. Collect the results from multiple GPUs and determine the fitness of each bicluster according to (\ref{eq:fitness}). 

\paragraph{Step 5: Update top-rank list.} Sort the population according to fitness. Try to add new individuals to the top-rank list by checking if they do not substantially overlap with records with higher fitness. If a bicluster is added, remove from the top-rank list all records that have lower fitness and substantially overlap with the bicluster. After all individuals in the population are checked, remove from the top-rank list the records that have the lowest fitness, until the required size of the top-rank list is reached. If the maximal number of iterations is not accomplished, go back to Step 2.

\paragraph{Step 6: Prepare biclusters.} Determine in parallel on each GPU the indices of rows that match each of the series of columns in the top-rank list. 

\paragraph{Step 7: Expansion of biclusters.} Expand the biclusters that have approximate and negative trends. Output the required number of biclusters (or all biclusters from the top-rank list).


\subsection{Pattern discovery on synthetic datasets.}
The performance of EBIC was evaluated on the benchmark of synthetic datasets from \citep{Wang2016} and compared to top biclustering methods: UniBic \citep{Wang2016}, OPSM \citep{BenDor2003}, QUBIC \citep{Li2009}, ISA \citep{Bergmann2003}, FABIA \citep{Hochreiter2010}, CPB \citep{Bozdag2009}, and BicSPAM \citep{Henriques2014}, as well as a newly published GPU-accelerated biclustering algorithm called Condition-dependent Correlation Subgroup (CCS) \citep{Bhattacharya2017}. The latter hasn't been benchmarked yet on the established collection of datasets, neither synthetic nor genomic. 

The test suite that was used to benchmark the algorithms contains three very popular biclustering problems: pattern discovery, biclusters overlap, and narrow biclusters detection. Recovery and relevance scores were determined using the Jaccard index \citep{Jaccard1901} from the BiBench package \citep{Eren2013}, specifically (\ref{eq:recovery}) and (\ref{eq:relevance}) :
\begin{eqnarray}
Recovery = \sum_{e \in expected} max_{f \in found} \frac{|e \cap f|}{|e \cup f|} \label{eq:recovery} \\
Relevance = \sum_{f \in found} max_{e \in expected} \frac{|e \cap f|}{|e \cup f|} \label{eq:relevance}
\end{eqnarray}

The first set of problems verifies the ability of the algorithm to identify six different data patterns, including trend-preserving, column-constant, row-constant, shift, scale, and shift-scale. The tests assess how accurately a biclustering algorithm detects three biclusters of size 15x15 implanted within a matrix of size 150x100, four biclusters of size 20x20 implanted within a matrix of size 200x150, and five biclusters of size 25x25 implanted within a matrix of size 300x200. Each problem consists of 5 different datasets for each of 6 patterns  -- which constitute 90 unit tests in total. The tests on overlapping patterns measure the ability of the algorithms to detect 5 biclusters of size 20x20 implanted within the matrix of size 200x150 that overlapped with each other by 0x0, 3x3, 6x6, and 9x9 elements -- 20 tests in total \citep{Wang2016}. Narrow biclusters are biclusters with 100 rows and 10--30 columns implanted within a large matrix of size 1000x100 -- 9 tests in total.  The tests determine whether biclustering methods are capable of discovering patterns that feature multiple rows but only a small number of columns \citep{Wang2016}. To show independence of the results, our method was run 10 times on all problems. Each time, a different seed served to initialize a pseudo-random number generator, which was used to initialize the population in the first iteration.

\subsection{Enrichment analysis on genomic datasets.}
The effectiveness of pattern discovery with EBIC was further evaluated on real-world gene expression datasets. For this purpose,  BiBench software and a benchmark of genetic datasets from Eren et al. \citep{Eren2013} were used. Details of the gene datasets used for the study are presented in Table  \ref{gds-datasets}. The same procedures of data acquisition, preprocessing, and analysis were followed.  Thus, datasets were downloaded using GEOquery \citep{Davis2007} and preprocessed using PCA imputation \citep{Stacklies2007}. After completing biclustering, a gene enrichment analysis of each bicluster was performed using the R package GOstats  \citep{Falcon2007}. Biclusters were considered significantly enriched if any of the p-values associated with a given GO term were lower than 0.05 after Benjamini-Hochberg correction \citep{Benjamini1995}. Assessment of the results was based on the proportion of enriched biclusters to all biclusters reported. Each algorithm was allowed to return no more than 100 biclusters per dataset. The number of biclusters found and the proportion of significantly enriched results were compared to the study by \citep{Wang2016} and are presented in Table \ref{gds-enriched-proportion}. EBIC was tested with two overlap ratios, 0.5 and 0.75.

\begin{table*}[ht]
\caption{Description of GDS datasets.}
\label{gds-datasets}
\begin{center}
\begin{tabular}{r|c|c|l}
\textbf{Dataset} &\textbf{Genes} &\textbf{Samples} &\textbf{Description} \\
\hline
GDS181 & 12626 &84 &Large-scale analysis of the human Transcriptome \\
\hline
GDS589 &8799 &122 &Multiple normal tissue gene expression across strains \\ 
\hline
GDS1406 &12488& 87& Brain regions of various inbred strains\\
\hline
GDS1451 &8799& 94& Toxicants effect on liver: pooled and individual sample comparison \\
\hline
GDS1490 &12488& 150& Neural tissue profiling \\
\hline
GDS2520 &12625& 44& Head and neck squamous cell carcinoma \\
\hline
GDS3715 &12626 &110& Insulin effect on skeletal muscle \\
\hline
GDS3716 &22283& 42& Breast cancer: histologically normal breast epithelium \\
\end{tabular}
\end{center}
\end{table*}

\section{Results}

The performance of EBIC was tested on both synthetic as well as real gene expression datasets. Synthetic benchmark from Wang et al. \citep{Wang2016} is available at \url{https://sourceforge.net/projects/unibic/files/data_result.zip}. For biological validation a well-established benchmark from Eren et al. was used \citep{Eren2013} with eight genetic datasets. The collection of datasets and the results of EBIC on both synthetic and genetic datasets could be found in Supplementary Data.

\subsection{Pattern discovery on synthetic datasets.}
\begin{figure*}[ht!]
\begin{center}
\includegraphics[width=0.9\textwidth]{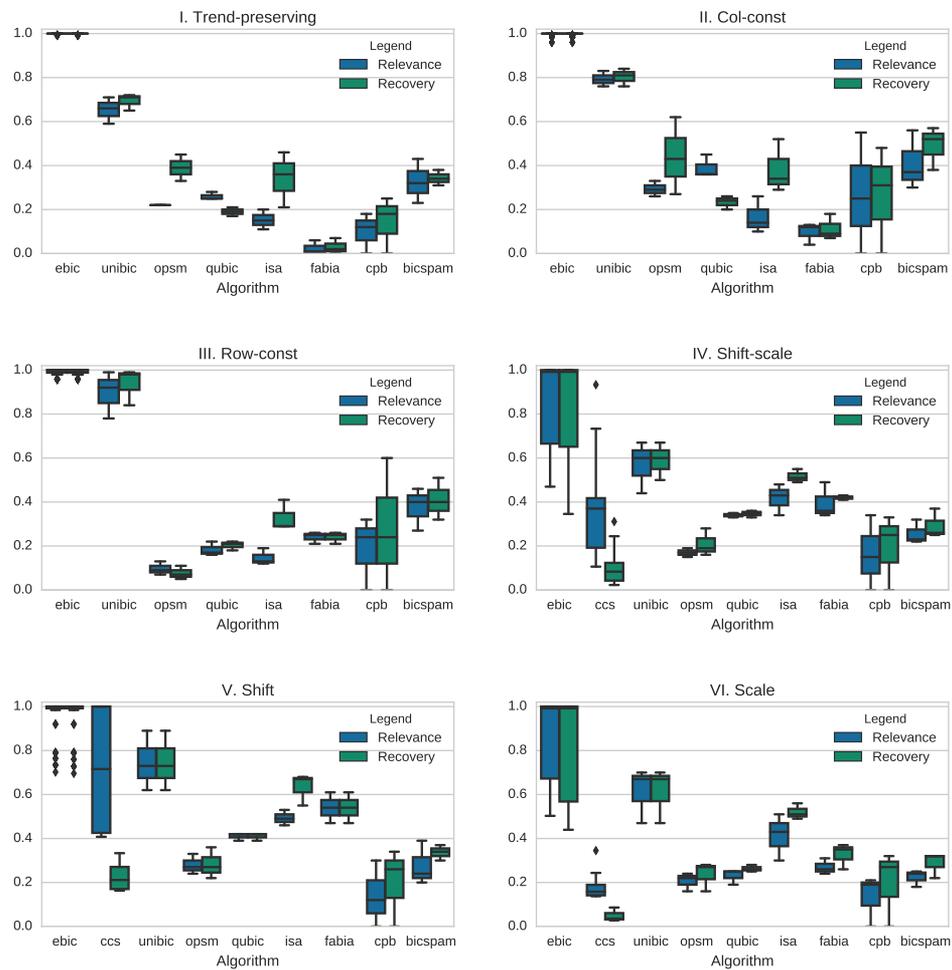}
\end{center}
\caption[Different types of biclusters.]{Comparison of the performance of biclustering algorithms on different types of patterns. Scores of the algorithms other than EBIC and CCS are quoted from \citep{Wang2016}.}
\label{types-performance}
\end{figure*}

For synthetic datasets EBIC was set to stop either after 20,000 iterations or when the number of tabu-list hits exceeded the size of the population. All parameters were set to their defaults. Columns of biclusters were allowed to overlap no more than 0.5, and the block-size for the CUDA kernel was set to 64. This took a reasonable amount of computation time (1-25 minutes on Intel Core i7-6950X CPU with GeForce GTX 1070 GPU). Comparison of the accuracy of EBIC with selected biclustering methods in terms of recovery and relevance is presented in Fig. \ref{types-performance}. CCS did not manage to return any result for trend-preserving, and row- and column-constant patterns, thus the method was excluded from the comparison. The CCS algorithm managed to present partial solutions for shift-, scale-, and shift-scale patterns only.

The average recovery and relevance scores of EBIC are better than those reported by any of the previous methods. This difference is especially visible in order-preserving and shift-scale problems, which are considered to be the most biologically meaningful \citep{Wang2016}. EBIC managed to detect all patterns perfectly for trend-preserving patterns, while other methods reached 70\% on average. The average relevance and recovery rate for shift-scale patterns were also much higher. As for scale and shift-scale patterns, EBIC attained high recovery/relevance scores across all tests (95.2\%/85.5\% for scale- and 94.2\%/84.5\% for shift-scale patterns), although scores for the worst-case scenarios were comparable to other methods (75.1\%/37.8\% and 72.0\%/46.7\%, respectively). EBIC may be the first biclustering algorithm capable of detecting all aforementioned patterns with over 90\% average recovery and relevance \citep{Pontes2015}. The recovery/relevance scores from multiple runs of the algorithm initialized with different random numbers did not differ statistically.

EBIC was also tested on the datasets provided by Bhattacharya et al. \citep{Bhattacharya2017} and detected biclusters with recovery and relevance scores over 95\%, whereas CCS reported those scores to vary from approximately 20\% to nearly 90\%.

\begin{figure*}[ht!]
\begin{center}
\includegraphics[width=0.55\textwidth]{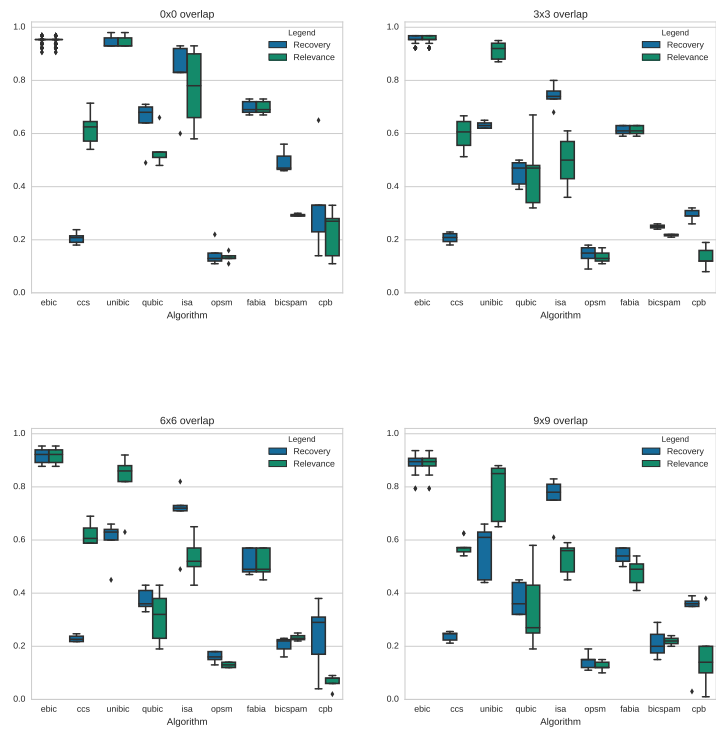}
\end{center}
\caption[The effect of overlap.]{Comparison of the performance of biclustering algorithms in scenarios with different levels of biclusters' overlap. Scores of the algorithms other than EBIC and CCS are quoted from \citep{Wang2016}.}
\label{overlap-performance}
\end{figure*}

\paragraph{Overlapping biclusters.}
The second set of tests compares the deterioration of the accuracy of biclustering algorithms in detecting trend-preserving biclusters that overlap with each other. This set of problems contains tests of 3 biclusters of size 20x20 that overlap with each other by 0x0, 3x3, 6x6, and 9x9 within a matrix of size 200x150. Each problem is represented by 5 dataset variants, resulting in up to 20 tests in total \citep{Wang2016}. The effect of the overlap on the recovery and relevance of different algorithms is presented in Fig. \ref{overlap-performance}. 

All biclustering methods tend to deteriorate if implanted biclusters start to significantly overlap with each other \citep{Wang2016}. The performance of EBIC also decreased when the higher level of overlap was considered, but the decrease was small. The algorithm was still able to maintain recovery and relevance scores close to 90\% on the average. The second-best method was UniBic, which deteriorated from around 90\% for non-overlapping biclusters down to 60\% recovery and 85\% relevance for the most overlapping structures.

\begin{figure*}[ht!]
\begin{center}
\includegraphics[width=\textwidth]{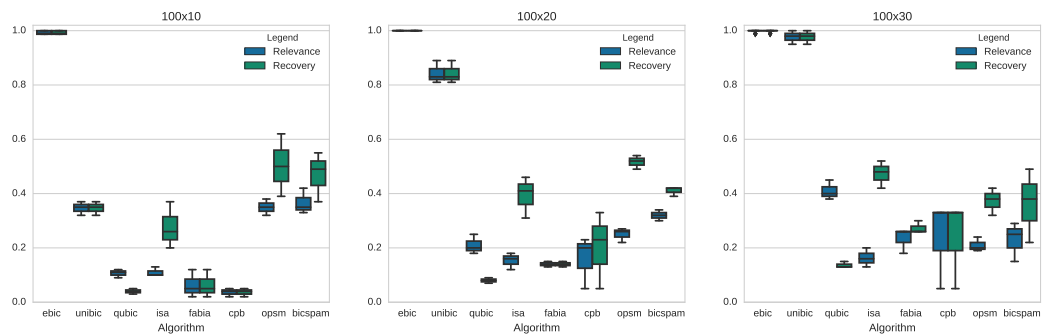}
\end{center}
\caption[Detection of narrow biclusters.]{Comparison of biclustering-algorithm performance in scenarios with narrow biclusters. The reference results are quoted from Wang et al.\citep{Wang2016}.}
\label{narrow-performance}
\end{figure*}

\paragraph{Narrow biclusters.}
The last phase of our benchmark considers the detection of narrow biclusters comprising 100 rows and 10/20/30 columns, which were implanted within the matrix of size 1000x100. Each scenario contains 3 variants, resulting in up to 9 tests in total. The results are presented in Fig. \ref{narrow-performance}.

In contrast to all other algorithms, EBIC managed much better in this task and discovered almost perfectly all implanted structures. For the narrowest biclusters, our algorithm was approximately twice as good as the second method dedicated to finding narrow biclusters (BicSPAM). UniBic was reported to have low accuracy in detecting narrow biclusters within the dataset. CCS did not manage to return any bicluster for every dataset in this test. 

\paragraph{Noise sensitivity.}
Noise sensitivity analysis of EBIC may be found in Supplementary Data. Tuning of EBIC parameters allows the method to be reasonably resistant to up to $N(0, 0.25)$ of normally distributed noise.

\paragraph{Summary.}
Our general conclusion is that EBIC is not only capable of detecting different types of patterns, but also different sizes of patterns (i.e., wide or narrow patterns) with very high accuracy. 


\subsection{Enrichment analysis on genomic datasets.}
For the genetic datasets, it was observed that the proportion of enriched biclusters obtained after approximately 5000 iterations highly depended on the dataset (see Supplementary Data). Further iterations either improved or worsened the proportion. EBIC was run for 5000 iterations, columns were allowed to overlap by 50\% or 75\%. The results of enrichment analyses are presented in Table \ref{gds-enriched-proportion}. 

Some memory management issues were encountered with CCS (both the sequential and parallel versions). The algorithm  was unable to detect biclusters in some of the genomic datasets and terminated prematurely with an error. After fixing a bug, the algorithm, even in parallel mode, proved to be extremely slow. Although the dataset was of reasonable size it took over 8 days of computation (on CPU+GPUs) to yield results for the most challenging genomic datasets (GDS 1451). In contrast EBIC needed less than 3 minutes to yield higher number of significantly enriched biclusters for this dataset.

EBIC generated the highest percentage of enriched biclusters. EBIC with a more restrictive overlap ratio (0.5) generated a higher percentage of significantly enriched biclusters (52.4\%) in comparison to any other method. The second best was CCS (43.5\%), which on the other hand generated much more significantly enriched biclusters. EBIC with less restrictive overlap (0.75) outperformed all the methods included in our study, both in terms of the number and percentage of significantly enriched biclusters. EBIC generated 20 significantly enriched biclusters more than the second-best method (323 vs 303 by CCS). More importantly, EBIC managed to find nearly 11\% more significantly enriched biclusters. This result is noticeable, considering that the difference between the second- and  third-best methods was only 2.7\%. In addition, the biclusters returned did not overlap substantially, from less than 4\% up to 31\%, depending on the dataset. The datasets as well as the results of biological validation of EBIC and CCS are available in Supplementary Data.

\begin{table}[ht!]
\caption{Significantly enriched biclusters found across all GDS datasets. Two overlap thresholds of EBIC are considered: 0.5 and 0.75. The scores of the algorithms other than EBIC and CCS are quoted from \citep{Wang2016}.}
\label{gds-enriched-proportion}
\begin{center}
\begin{tabular}{r|c|r}
Algorithm &Found &Enriched \\
\hline
\textbf{EBIC, 0.75} &\textbf{589}&\textbf{323 (54.8\%)} \\
\hline
EBIC, 0.5 &145&76 (52.4\%) \\
\hline
CCS & 691 & 303 (43.8\%)\\
\hline
UniBic &151 &62 (41.1\%)\\
\hline
OPSM &163 &48 (29.5\%)\\
\hline
QUBIC &91& 34 (37.4\%)\\
\hline
ISA &217 &71 (32.7\%)\\
\hline
FABIA &80 &22 (27.5\%)\\
\hline
CPB &96 &34 (35.4\%)\\
\end{tabular}
\end{center}
\end{table}

We reinspected the results of the two best methods (EBIC-0.75 and CCS) after applying a filtering proposed by Prelic et al. \citep{Prelic2006} and implemented in Eren et al. \citep{Eren2013}. The procedure removed biclusters that overlap with the others by over 25\%. After filtering, 296 out of 589 biclusters for EBIC remained, out of which 122 were found to be significantly enriched (41.2\%). For CCS, 332 out of 619 remained and only 113 were marked as significantly enriched (34.0\%). We eschewed testing the other methods, as their number of significantly enriched biclusters before the filtering procedure was even applied was lower than the one for EBIC or CCS after applying the procedure.

\subsection{Scalability of the algorithm.}

\begin{figure}[ht!]
\begin{center}
\includegraphics[width=0.5\textwidth]{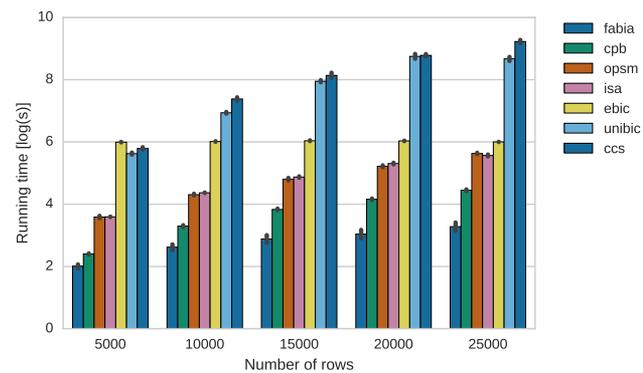}
\end{center}
\caption[Run times of the algorithms.]{Comparison of running time of the algorithms on datasets with 100 columns and varying numbers of rows.}
\label{running_times}
\end{figure}

In order to assess the scalability of the methods, five datasets with 100 columns and different numbers of rows ranging between 5000 and 25000 were generated. Times were averaged based on five runs of the methods on each of the datasets with their default parameters. The algorithms were allowed to yield up to 100 biclusters. All tests were performed on a machine with an Intel Core$^{TM}$ i7-6950X CPU and 64GB of RAM.  Comparison of run times in logarithmic scale is presented in Fig. \ref{running_times}. Starting with 10,000 rows, EBIC began to run faster than both CCS and UniBic, the most precise methods so far. For problems with 25,000 rows, EBIC was over 12 times faster than UniBic and over 20 times faster than CCS. With increasing data size, the running times of EBIC have started to be comparable with ones from OPSM and ISA. The actual performance of EBIC for larger datasets on multiple GPUs requires further investigation. 

A complexity analysis of EBIC can be found in Supplementary Data.


\section{Discussion}
EBIC is one of the very few parallel biclustering methods dedicated for multi-GPU environments. In comparison with state-of-the-art algorithms EBIC exhibited a number of advantages: (1) EBIC outperformed the state-of-the-art biclustering algorithms on established synthetic datasets. EBIC was the only algorithm to discover each of six types of major genetic patterns in synthetic datasets with over 95\% average accuracy and the only one to maintain over 90\% accuracy on narrow and overlapping biclusters. 
(2) EBIC found over 11\% more significantly enriched biclusters than the second-best method (CCS) on a benchmark of 8 genomic datasets (over 7\% more after removing overlapping biclusters). (3) EBIC yielded far more significantly enriched biclusters than any of the methods (even after removing overlapping biclusters). (4) EBIC proved to be over 12 times faster than any of the most accurate methods (CCS or UniBic) on the largest datasets. 

We would like to formulate the requirements for the next-generation of biclustering methods. Such algorithms are expected to meet the following criteria: (1) be capable of discovering the six major types of biclusters discussed above with high accuracy (over 75\% on average); (2) be capable of handling overlapping, narrow, and approximate patterns with similar accuracy; (3) provide meaningful solutions for both synthetic and real datasets; (4) be scalable. In contrast to other methods described in this paper, EBIC with its average accuracy exceeding 90\% certainly meets these requirements and could be called a next-generation biclustering method. 

EBIC has certain limitations. First, the closer the overlap threshold to 0, EBIC may no longer be able to capture different series that are present within the same columns. Instead, this series of columns which is represented by the largest number of rows will incorporate all other permutations. The reason for this is construction of top-rank list. For performance purposes, the list uses intersection of columns as the merging criterion, what makes the actual order of columns within the series irrelevant. A full overlap of biclusters within the top-rank list is possible, but discouraged. Secondly, application of EBIC to datasets that have fewer than 20 columns is discouraged. In this case an exhaustive search guarantees discovery of all meaningful patterns in a much shorter time. Thirdly, the overlap degree of biclusters for a dataset requires verification. Tuning the parameters of the method may decrease the level of overlaps. A more restrictive overlap threshold (0.5) allows the algorithm to detect fewer biclusters with less overlapping columns, while a less restrictive overlap threshold (0.75) returns more biclusters at the cost of their overlap. The degree of biclusters' overlap cannot be directly controlled in EBIC.
    
The guidelines for the exact number of iterations to run EBIC, as well as the optimal level of overlap on biclusters in the top-list, need to be empirically defined. EBIC scores do not seem to improve with every iteration. The accuracy of pattern detection generally improves over time for synthetic datasets, but this did not hold for real genomic datasets. The highest proportion of significantly enriched biclusters oscillated or even slightly deteriorated for real-world genetic datasets after 100 iterations. For all genomic datasets, EBIC was stopped after 5,000 iterations, as it seemed to be a reasonable compromise between the percentage of enriched results and run time. Additional study on the influence of the size of the input matrix on the number of required iterations is needed.

Our initial tests using larger volumes of data indicate that the algorithm supports datasets of up to 60k rows per GPU. Full scalability of EBIC and preparing the algorithm for big data challenges requires more work.

\section{Conclusions}

EBIC is anticipated to become a reference method for future studies in biclustering. EBIC may also prove beneficial in other domains beyond genomics. The method may improve pattern detection in multiple other fields (e.g. medicine, applied informatics, economics, biology, or chemistry) in which biclustering has been previously successfully applied. Extensive AI method development is necessary to fully realize the potential of AI for solving the most challenging big data problems.

\section*{Authors' contribution}
P.O. conceived the study, designed and implemented the algorithm. P.O. and J.H.M. performed the analysis. M.S. and X.H. consulted the project, analyzed the results and participated in writing the manuscript. J.H.M. oversaw the project.

\section*{Acknowledgements}
This research was supported in part by PL-Grid Infrastructure and by National Institutes of Health grants LM012601, TR001263, and ES013508.

\bibliographystyle{natbib}
\bibliography{bibliography}

\end{document}